# A Review on the Applications of Machine Learning for Tinnitus Diagnosis Using EEG Signals

Farzaneh Ramezani[1] , Hamidreza Bolhasani[2+]

**Abstract-Tinnitus is a prevalent hearing disorder that can be caused by various factors such as age, hearing loss, exposure to loud noises, ear infections or tumors, certain medications, head or neck injuries, and psychological conditions like anxiety and depression. While not every patient requires medical attention, about 20% of sufferers seek clinical intervention. Early diagnosis is crucial for effective treatment. New developments have been made in tinnitus detection to aid in early detection of this illness. Over the past few years, there has been a notable growth in the usage of electroencephalography (EEG) to study variations in oscillatory brain activity related to tinnitus. However, the results obtained from numerous studies vary greatly, leading to conflicting conclusions. Currently, clinicians rely solely on their expertise to identify individuals with tinnitus. Researchers in this field have incorporated various data modalities and machine-learning techniques to aid clinicians in identifying tinnitus characteristics and classifying people with tinnitus. The purpose of writing this article is to review articles that focus on using machine learning (ML) to identify or predict tinnitus patients using EEG signals as input data. We have evaluated 11 articles published between 2016 and 2023 using a systematic literature review (SLR) method. This article arranges perfect summaries of all the research reviewed and compares the significant aspects of each. Additionally, we performed statistical analyses to gain a deeper comprehension of the most recent research in this area. Almost all of the reviewed articles followed a five-step procedure to achieve the goal of tinnitus. Disclosure. Finally, we discuss the open affairs and challenges in this method of tinnitus recognition or prediction and suggest future directions for research.**

[1] Department of Psychology, University of Tabriz, Tabriz, Iran
[2] Department of Computer Engineering, Islamic Azad University Science and Research Branch, Tehran, Iran.
[+] Corresponding Author: hamidreza.bolhasani@srbiau.ac.ir

## 1. Introduction

Tinnitus is a sort of phantom perception defective by neural activities associated with the clutter of the auditory system [1] and characterized by hearing undesirable sounds that are not present evidently [2]. Many people encounter a stable noise in their ears, further reported as a whistling or ringing sound in the ears [3]. Tinnitus is one of the three most prevalent clinical issues in otology. It is a common disease, affecting about 10-15% of the world's society and up to 33% of the elderly [1] are pretentious by tinnitus [4], and 10–20% of them indicate that tinnitus interrupts their daily life [5]. Tinnitus can cause insomnia, damaged cognitive capability, and difficulties in mental concentration, Serious people even appear anxious or depressed, disturbing the patient’s routine life [4]. Despite its extensive prevalence, the pathogenesis of tinnitus is uncertain, in most of these occasions, tinnitus is a personal perception that can only be comprehended by the influenced person [6], and clinical assessment and inspection are mostly based on the patient’s medical history, signs, auditory system trial, evaluation scale, psychoacoustic analysis, and absence of more objective recognition and assessment methods [7]. Spacious trials and research investigating the source of tinnitus have direct to the extensively embraced belief that tinnitus may be operated by momentarily anxious and annoying conditions, but turned into an indefinite sign by an unfamiliar mechanism in principle auditory pathways [8], [9]. To restore the loudness of tinnitus, activation of the auditory cortex in the resting state has been suggested [10]. Aspects of tinnitus such as distress correspond to the activation of non-auditory regions such as the frontal region [11]. Neural conversions linked with tinnitus contain conversions in the level of spontaneous neural activity, attendant neural switching, and reorganization of cortical tonotopic maps [12].

Different functional imaging methods have been used to localize brain areas related to tinnitus [12, 13], and the most popular method is electroencephalography (EEG) [14]. EEG is a medical procedure that examines the electrical activity of the scalp generated by brain structures. Macroscopic searches of the superficial layer of the brain below [15], [16], [17], [18] have been shown to show abnormalities in frontal, temporal, cingulate, parietal, and other areas of the brain. EEG frequency bands have been extensively researched [19], [20] due to the non-invasive and real-time reflection of the brain condition [21], EEG is spacious used in middle mechanism prediction, diagnosis, and treatment. Tinnitus. Compared to other neuroimaging techniques such as functional magnetic resonance imaging (fMRI) or magnetoencephalography (MEG) it has excellent temporal resolution and cost-effectiveness. [22].

The latest research has used ML to bring down prop on adept and diminish the impact of personal cues in the tinnitus-diagnosis procedure [23, 24, 25]. The most obvious and ordinarily artificial technology used to advance tinnitus therapies is ML [26]. Machine learning models do not allow computers to make complex abstractions of simple propositions [27]. Nevertheless, most machine-learning EEG studies on tinnitus utilized resting-state EEG and focused on model performance [25, 28, 29]. However, caused of the intricate central mechanism of tinnitus and countless influencing factors, many previous researches have not achieved general consistent outcomes. Accordingly, many intentions have been made to diagnose tinnitus relying on the EEG signals by utilizing ML methods [30, 31].

Our new diagnostic tool, based on machine learning, uses discriminative features to accurately identify patients suffering from tinnitus [32]. It was productively exhibited that machine learning presents’ a methodology with numerous possibilities for predictive precision, proficiency, sanity, and Possibility of generalizing the question [33].

In recent times, there has been a rise in the availability of ML and Deep learning (DL) in predicting and detecting tinnitus with the aid of EEG signals. The original target of this article is to manner a systematic literature review (SLR) of papers that focus on the application of machine learning in detecting or predicting tinnitus using EEG signals. Our SLR survey is the first of its kind in this area of research as no other systematic literature review has been done on this topic before.

This review makes several significant contributions which are outlined below:

- Create a comprehensive taxonomy chart that covers all the machine-learning algorithms that are currently being used
- Concisely summarizing researched material and comparing main aspects
- Propose a five-step pattern, including information for each stage, to analyze all reviewed papers
- Identify any existing issues and areas for improvement to better address them in the future

The review is structured into several sections, each focusing on a different outlook of the research. Section 2 covers pertaining works, while Section 3 explains the research manner adopted. In section 4, you can find the designed taxonomy and provided summaries. Section 5 offers a discourse and collation of the studied papers, and finally, section 6 presents the conclusion.

## 2. Review of Related Studies

We could not find any review articles that used both EEG signals and machine learning algorithms to examine tinnitus. In this section, we will analyze two nearly identical articles.

In a review conducted by Fan S et al. [34] various articles were analyzed regarding the use of EEG signals and classifiers in detecting tinnitus. The review focused on objective methods for evaluating tinnitus, examining potential waveform components recorded through procedures like brain stem auditory proliferation, event-related potential, crack intuition, and EEG. The review discusses the current state of research, the constraints of these methods, ongoing rivals in the field, and potential future research directions. Recent findings suggest electrophysiological approaches may be able to detect tinnitus by analyzing neural activity in the auditory pathway of affected individuals. This analysis may help identify parameters for these methods and further probe the eventuality of the utilization of electrophysiological methods for objective tinnitus demodulation. The findings from various studies may vary due to several factors like the method of tinnitus induction in animal trials, the cause and categorization of tinnitus in humans, the accuracy of tools and techniques used, demographic traits of test subjects, and more. Hence, there are no universal standards in this area.

In research done by Durai M et al. [35] the individual responses to tinnitus and its psychological effects were examined about three types of masking techniques. These techniques were energetic masking informational masking, and a combination of both. The study followed 11 participants who were suffering from chronic tinnitus over a cycle of 12 months. Each participant utilized each masking technique for 3 months with a 1-month washout period in between. The EEG data was computationally modeled using NeuCube, a Spiking Neural Network (SNN) architecture inspired by the human brain. The NeuCube framework was specifically designed for this study to map, learn, visualize, and classify brain activity patterns. The SNN model was used to study the relationship between EEG and clinically significant changes in the TFI. EEG and psychoacoustic assessments were conducted at the beginning and after three months of each masking sound treatment. The EEG data was computationally modeled using the NeuCube framework, a brain-inspired SNN architecture premeditated for this study. The framework enabled mapping, learning, visualization, and classification of brain activity templates. The SNN model demonstrated that EEG was closely linked to clinically significant changes in the Tinnitus Functional Index (TFI). In addition, the SNN framework accurately predicted sound therapy responders (93% accuracy) and non-responders (100% accuracy) utilizing baseline EEG recordings. The compound of strenuous and data masking was found to be a more effective treatment sound for a larger number of individuals than other sounds used in the study. However, the results are preparatory and need confirmation in larger and independent instances.

Table 1. Major traits of studied research included tinnitus diagnosis and prediction using ML and EEG

| Reference | Publish year | Major field |
|---|---|---|
| Shuwen Fan et al. [34] | 2022 | Using EEG signals for tinnitus diagnosis |
| Durai M et al. [35] | 2020 | Using an SNN model for the prediction of tinnitus masking benefit |

## 3. Research Method

In September 2023, we conducted a systematic literature search on PubMed, Web of Science, and Scopus, following the Meta-analysis of Observational Studies in Epidemiology guidelines (MOOSE), as outlined in the Supplementary Materials (Stroup et al., 2000). The search included a range of keywords related to tinnitus such as "tinnitus prediction," "tinnitus detection," "tinnitus electroencephalographic evaluation," "tinnitus based on EEG," "tinnitus and deep learning," and "tinnitus and machine learning." We also included relevant studies that appeared in the reference lists of the selected articles. There were no language restrictions in our search.

The following Analytical Questions will be completely answered in this SLR paper.

AQ1: Which machine learning algorithms have been used to diagnose or predict tinnitus?
AQ2: Which machine learning methods are most preferable to achieve our purpose?
AQ3: What are the primary methods utilized for extracting features from EEG signals?
AQ4: What are the future and open works relevant to tinnitus diagnosis or prediction?

All selected papers were recognized by the authors and appraised against inclusion and exclusion criteria.

The rudimentary search resulted in 33 articles. The number of repeated cases was 28 studies. After reviewing the abstracts of these articles, 20 studies were selected to read the complete text, and 8 studies were cut off due to not meeting the inclusion criteria. Finally, a whole of 11 essays were chosen (see Figure 1 for the selection process).

Fig 1 illustrates the evaluation process for the selected studies. Ultimately, 11 papers were chosen for consideration.
Inclusion essentials are formulated as comprising perusals that have:

- broadcasted between 2018 and 2023
- Focused on accord machine learning to diagnose or predict tinnitus using EEG signals

Exclusion elements are compared in perusals that have:

- Not profile in ISI
- Not written in English

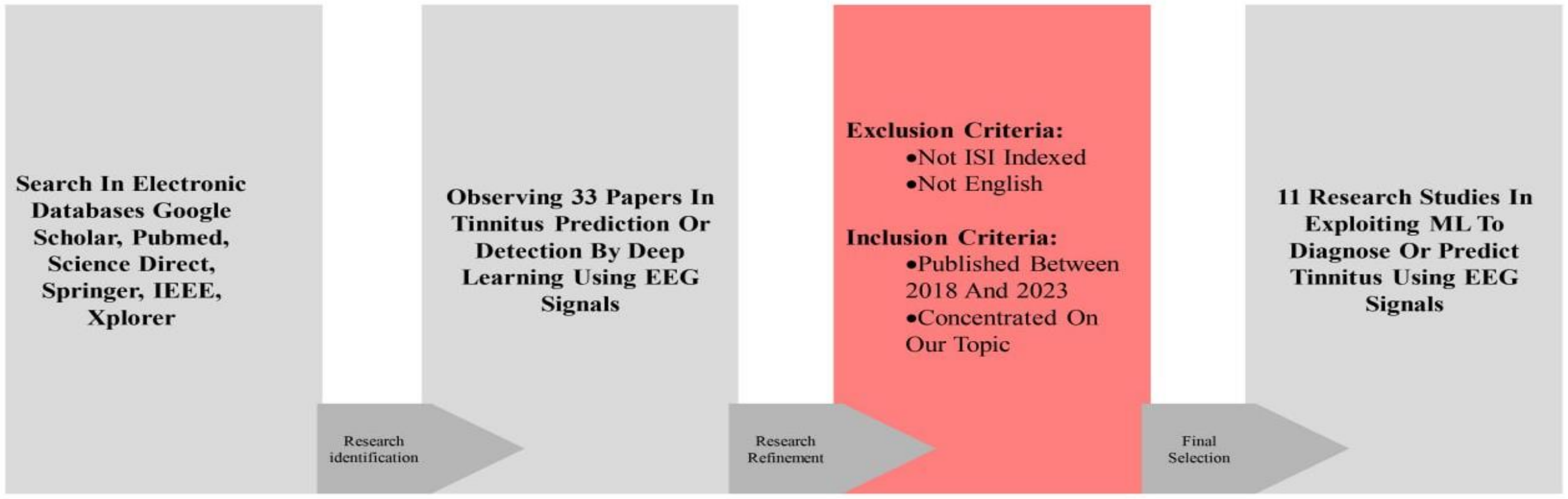

Fig. 1. The paper selection and appraisement flow chart

## 4. Structure of machine learning algorithms for tinnitus diagnosis using EEG signals

This section provides a summary of research papers that have used ML algorithms for diagnosing or predicting tinnitus based on EEG signals. These papers have followed the systematic literature review process and adhere to the rules mentioned in Section 3. Fig.2 shows an encyclopedic taxonomy of the ML and DL used for tinnitus diagnosis and prediction. These models fall into four main categories: SVM models, CNN models, composition models (combined models of both support vector machine and Convolutional Neural Networks), and Other Algorithms. Since machine learning algorithms other than CNN and SVM were less widespread and minor popular across the selected papers, they have been classified as "Other Algorithms" clusters. Therefore, the charts presented in this article have been checked according to the algorithms used in all the articles. The study is divided into two subsections 4.1 and 4.2 respectively. In subsection 4.3, a complete analysis of the articles and a comparison between them is given. The main concepts such as the main context, positive and negative points, and contributions are presented in Tables 2 and 3. It was observed that all articles follow a general five-step pattern to differentiate between tinnitus and normal cases. Subsection 4.3 analyzes these five key stages, and Fig. 3 displays them. Subsections 4.3.1 to 4.3.5 present data pertinent to each of these five steps in tables numbered 4 to 9, providing an occasion for analogy among the inquired articles in terms of several techniques and procedures employed in every phase.

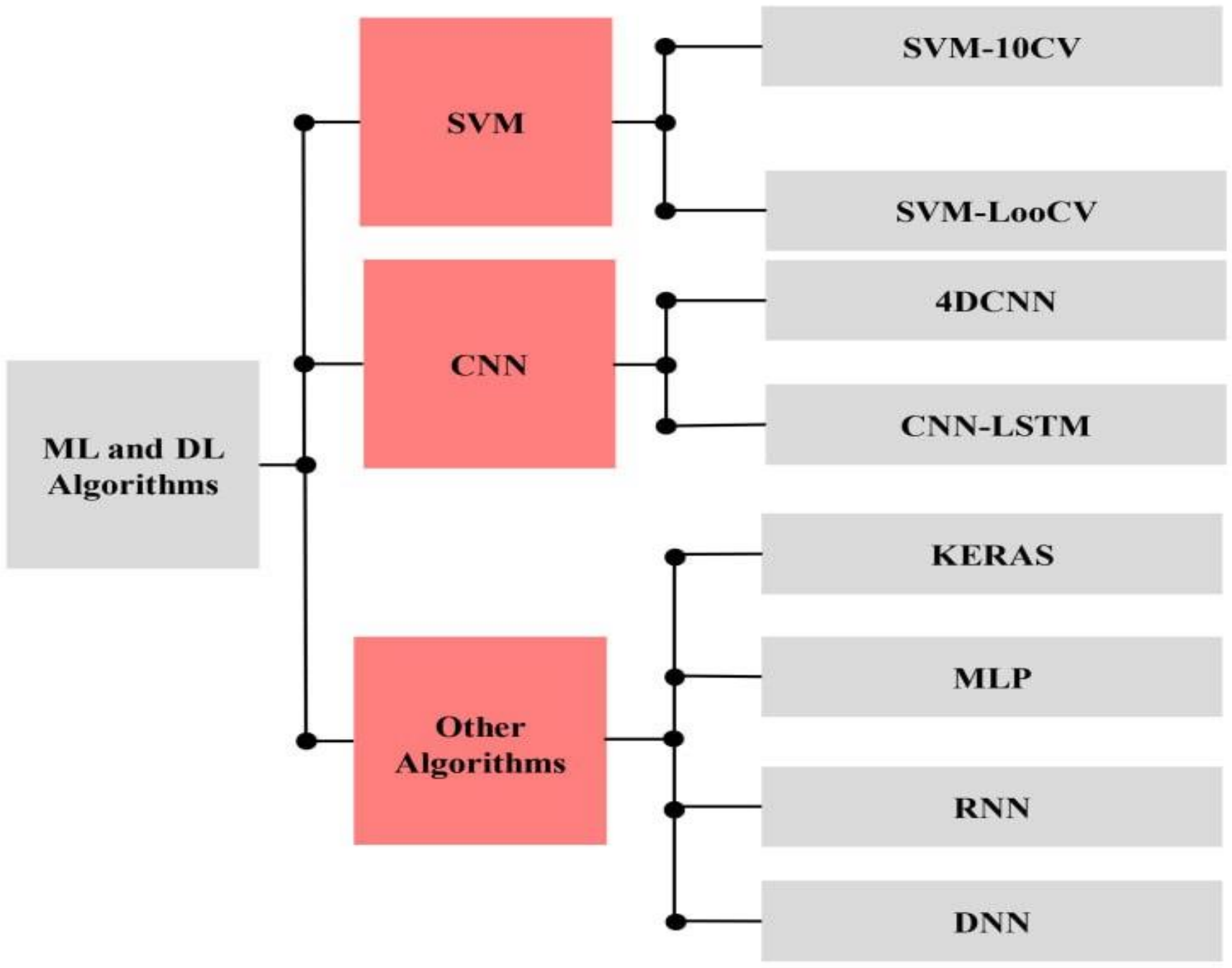

Fig.2. The assortment of ML models for tinnitus recognition using EEG signals

### 4.1 Machine learning methods for tinnitus recognition using EEG signals

By the structure of SLR, this section aims to purvey a concise, yet comprehensive introduction to all the articles examined. Its purpose is to familiarize future researchers with prior studies conducted in the context of this review's subject. The section strives to cover all chief parts and visage of the essays in the figure of abstracts, including information about the dataset used, structure of the proposed ML model, methods of data provision, privilege, and disadvantages of the pursuing approach, as well as the achievements and contributions of each study.

Mohagheghian F et al. [36] used the Weighted Phase Lag Index (WPLI) to analyze the data across different frequency bands ranging from 2-44 Hz. The classification of the data was done using SVM, and graph theoretical measures were used to extract classification features from the connectivity matrices. To ensure minimal redundancy, the researchers used feature ranking and Principal Component Analysis (PCA) which generates mutually uncorrelated features. All the flairs were normalized using linear scaling, and feature ranking was done using an entropy-based approach. To achieve the most suitable classification proficiency with a reasonable running time, the top 60 features were selected. Finally, the first few principal components of PCA were used to create a new feature vector. Non-linear SVM classification was completed using RBF kernel. To evaluate the classification performance, the Leave-One-Out Cross-Validation (LOOCV) method was employed due to the limited number of instances. The study produced encouraging results with high accuracy, sensitivity, and specificity across all frequency bands, especially in the beta2 band. However, the main limitation of this research was the small sample size of participants.

Wang C D et al. [37] expanded a deep neural network model utilizing the MECRL framework on a dataset they created. The model accurately distinguishes between tinnitus patients and healthful individuals. The researchers compared the effectiveness of the MECRL frame with traditional ML methods that used flair engineering and multiple EEG-based deep learning methods, including v-SVM, MLP, EEGNet, SiameseAE, SMeta-SAE, and 4D-CNN. The study showed that methods using deep features extracted from raw EEG data generally outperformed methods using handcrafted features as input. However, EEGNet and 4D-CNN were trained under a simple supervised learning model, which made it difficult for the models to accurately classify data from unseen subjects. On the other hand, MECRL deconstructed the complex EEG data into a dynasty of self-monitored learning functions that were advanced and semantically complementary, allowing the model to map EEG segments into effective value and discriminate semantic representations. This study helped better understand the relationship between electrophysiology and pathophysiological shifts of tinnitus and provided a new DL approach to laterality, hearing state, and tinnitus intensity on brain networks that were not more prospect in the present etude. However, the research is limited by the small sample size.

A study conducted by Piarulli A. [38] has suggested that EEG when combined with advanced classification techniques, can accurately identify biomarkers that differentiate levels of distress in tinnitus patients. The study utilized two classifiers - one for identifying tinnitus and the other for identifying distress levels - with average accuracies of 89% and 84%, respectively. The features used in each classifier had minimal overlap. To select the subset of traits that performed the best in differentiating between the two groups, the researchers used a feature selection algorithm that utilizes neighborhood portion analysis. This algorithm tracks features that greatly increase the prediction accuracy of the classifier and has been run 1000 times to ensure stability. Features with the highest discrimination rates were retained for better analysis. The SVM algorithm was used to test the performance of the selected features for classification. The SVMs were run with 5-fold cross-validation 1000 times to ensure constant reputation resolution. However, the researchers suggest that DL classification methods with neural networks may be more adequate than SVMs for certain sorts of EEG analysis.

Jianbiao M et al. [39] revealed that entropy can be used to measure the level of chaos in the brain. The chaotic properties of the torpid situation EEG signal can help diagnose tinnitus in its early stages. EJ data was divided into training and testing. Then, SVMs were selected using the heuristic search method. SVMs increase the number of edge regions and greatly reduce testing errors. The selected support vector determines the decision and computational effort and removes the redundant features. During the test, the flair set (N) was divided into ten equal subsets. The remaining 9n/10 portion of the feature set was used to train and test the residual features (n/10) 10 times. The average resolution was used as a dependable classification efficiency measure to compute its accuracy (Acc), recall (Re), precision (Pre), and F1. One of the limitations of the study is the small sample size. Additionally, it did not inquire about pre-test medication usage for tinnitus.

Resting-state EEG signals were recorded in tinnitus patients with varying locations of tinnitus by Lie Z et al. [40] The study aimed to classify the possible tinnitus sources based on the selected traits. Six different features were exploited from the EEG signals, including four connectivity features (PLV, PLI, and PCC) and two time-frequency domain traits. Four ML algorithms were exploited to analyze the features. The PCC trait sets, SVM, and MLP algorithms achieved the highest classification accuracy, with average scores of 99.42% and 99.1% respectively. The PLV feature set also yielded good results. The MLP algorithm was the fastest, taking only 4.2 seconds, which makes it suitable for real-time diagnosis. However, this study only focused on tinnitus patients without hearing loss. Further experiments are essential to vouch for the effectiveness of connectivity traits in individualized tinnitus patients from healthy individuals with hearing loss.

Sun Z R et al. [41] achieved a significant result in secretive tinnitus patients from healthy individuals by merging different views of EEG data using the Support Vector Machine classifier. The method resulted in a precision, summon, and F1 score of 99.72%, 98.97%, and 99.34%, respectively. This approach is both effective and objective and warrants further research of ML methods to predict the efficiency of tinnitus interposition ground on the EEG response of tinnitus individuals. The EEG signals were amplified and band-pass-filtered to a frequency range of 0.5 to 80 Hz. The data was commissioned versus 56 and 107 mastoid electrodes bilaterally. To ensure accuracy, any episodic artifacts such as eye blinks, teeth clenching, body movement, or ECG artifacts were removed from the EEG waves using portable artifact rejection and ICA. The halting data was then divided into at least 20 2-second epochs. After that, average Fourier cross-spectral matrices were adjusted for the concerned frequency bands. This ensures that the data is reliable and accurate. The SVM algorithm targets to find the proper "hyperplane" to discrete the data attached to various classes. After capturing the features matrix of the dataset in the latent intact region by multi-view whole area learning, the SVM method (with RBF kernel) is applied to the trait matrix to carry out the manner of classification. However, the lack of sufficient input data is one of the main problems of this research.

Z S et al. [42] used the wavelet transform to exploit four frequency components (delta, theta, alpha, and beta) from EEG signals. Power spectrum entropy was then calculated for each frequency band, and the DNN model was trained with the eigenvalues. DNN consists of a 4-dimensional input layer, two hidden layers with 8 neurons, a RELU activation function, and an output layer. The EEG signals of 26 samples were taken from their left middle temporal lobe and sorted by CNN, RNN, and DNN models. Among the models, the DNN model had the highest classification accuracy (92%). This study investigated the relationship between EEG signals at rest and tinnitus and introduced DNN as the best model for tinnitus auxiliary diagnosis. The CNN network added eight 1x1 small convolutional cores, while the RNN model used LSTM to untangle the issue of gradient vanishing in traditional RNN networks. The RNN model had a structure of 4x8x8x1 with two LSTM hidden layers, each with 8 neurons. In this model, the first hidden LSTM layer utilized the predicted value of the whole sequence as feedback to the next layer. On the other hand, the second hidden layer of LSTM did not employ the predicted worth of the progression as feedback. Nonetheless, the model's main drawback is attributed to the limited amount of input data it can handle.

Hong ES et al. [32] conducted the SVM method, and the alpha-band time strings of the whole EEG trial at the Pz electrode were utilized as a classifier input trait because the alpha activity is universally dominant in the parietal area. To check the detection of the alpha band in SVM decoding efficiency, AUC was also calculated in the state of alpha band removal, through the band-stop filter, from the input signals. Using this trait, we handled an SVM with the radial-basis operation as a kernel. The arrange and kernel parameters were chosen using grid search. To understand the contribution of each frequency band in the model training, we evaluated the performance of EEGNet on narrowband data and analyzed the characteristics of the convolutional layer filters trained on wideband data. The EEGNet model was trained on both broadband data and EEG signals from individual frequency bands, which were delta, theta, alpha, beta, gamma, and broadband. The EEG data underwent filtering in these frequency bands. They found that, during an eccentric task, tinnitus patients could be noticed with a region under the curve of 0.886 via EEGNet using EEG signals. Also, applying broadband EEG signals, and analytics of EEGNet convolutional kernel traits maps disclose that alpha pursuit might play a vital figure in noticing people with tinnitus. However, the lack of sufficient input data was the main flaw of this research.

In a study conducted by Vanneste S et al. [43], a region of interest (ROI)-based approach was used to identify brain regions involved in the pathophysiology of tinnitus. After conducting a meta-analysis, the researchers were able to select the ROI. Using beta, theta, and gamma bands, the SVM model was able to differentiate between individuals with tinnitus and healthy individuals with an accuracy of 87.71%, compared to a random model that was only 53.30% accurate. It is worth noting that the gamma frequency for the dorsal anterior cingulate cortex was included in the study. Among the problems of this study, it can be mentioned that dealing with several neurological diseases at the same time has reduced the accuracy of the study on tinnitus patients.

### 4.1.1 Resolution of reviewed articles centralized on tinnitus diagnosis

Section 4.1's essays are compared and contrasted in Table 2 to highlight differences and backgrounds.

Table 2. Storing of essays concentrated on tinnitus assessment by ML

| Reference | Publish Year | Best Method | Privilege | Disadvantage | Contribution |
|---|---|---|---|---|---|
| Piarulli A et al. [38] | Sep-23 | SVM | • Analysis features with higher discriminatory capacity | DL classification methods with neural networks may be more suitable than SVMs | EEG-based framework |
| Wang CD et al. [37] | Jul-23 | MECRL method combined with vSVM, EEGNet, 4D-CNN, SiameseAE, MLP, SMeta-SAE, 4D-CNN | • provides a new DL method (MECRL)<br>• Multiple layer by layer progressive learning tasks | Small input data | EEG-based framework |
| Hong ES et al. [32] | Apr-2023 | Hybrid EEGNet model with alpha band | • Made comparison between EEGNet and SVM | limited sample size | Adaptation CNN for EEG signal processing |
| Jianbiao M et al. [39] | Jul-22 | SVM combined with optimal feature combinations | • combines time-frequency domain and non-linear power analysis | More data and other changes in the structure to check the low accuracy | EEG-based framework |
| Sun ZR et al. [42] | Jul-22 | SVM method with RBF kennel | • Characterization of EEG signals by feature extraction in a hidden intact space | Small input data | Machine learning based on EEG data |
| Li Z et al. [40] | Jan-22 | EEG signals combined with four machine learning algorithms | • Using binding properties (PCC and PLV) as biomarkers | Small sample size | Combination of deep learning and machine learning |
| Z S et al. [43] | Aug-21 | DNN, CNN and RNN models | • Distinction between DNN layers | Small data size | EEG signals Based on WT and DNN |
| Mohaghegh hian F et al. [36] | 2019 | • Weighted Phase Lag Index (WPLI)<br>• SVM | • Measuring brain networks based on EEG | Small number of subjects | SVM framework |
| | | | functional connectivity | | |
| Vanneste S et al. [44] | 2018 | SVM learning model | • Comparison TPR, FPR and ROC | Small sample size | Thalamocortical dysrhythmia (TCD) |

### 4.2 ML and DL manners for tinnitus prediction using EEG signals

Liu Z et al. [44] present a new, efficient method for discerning tinnitus from a healthy condition using EEG-based tinnitus neurofeedback. The proposed method includes a trend descriptor, which is a feature extractor that reduces the impact of electrode noise on EEG signals. Additionally, it incorporates a siamese encoder-decoder network that learns precise alignment and high-modality transferable mappings across subjects and EEG signal channels in a supervised system. The test results show that this method outperforms state-of-the-art algorithms and achieves a resolution of 91.67%-94.44% when predicting tinnitus and control subjects in a subject-independent setting. The erosion studies on mixed subjects and parameters show the stable performance of the proposed method, which utilizes v-SVM for predicting class labels based on orientation descriptors and autoencoder attorney. The method was compared with several competitive methods and outperformed them in both experiments, demonstrating its robustness and a stronger capability to catch the subject variance.

Allgaier J et al. [45] used the Keras model for the machine learning process and neural network architecture. To ensure the correctness of the model choice and prevent accidental changes during the architecture conversions, random seeds such as Numpy, TensorFlow, and the OS environment were set. This model has 5 dense layers, an elimination layer, and a flat layer. A sigmoid layer was set up because it was a binary classification task. The neural network was educated for one period at a time and the number 32 was chosen as the batch size. A larger batch, unlike a smaller batch, leads to memory problems the model was trained using binary cross entropy and RMSprop optimizer with a learning rate of 0.001. Using this research, EEG signals can be followed to extract features related to tinnitus and increase the prediction accuracy. However, this study had two major limitations. First, the EEG data for the trial is insufficient and it is unclear whether the trial participants are representative of the broad spectrum of tinnitus patients.

#### 4.2.1 Resolution of reviewed papers concentrated on tinnitus prediction

Section 4.2 articles imply solidarity as Table 3 shows major opinions.

Table 3. Detail essay information focusing on tinnitus prediction by ML

| Ref | Year of Publication | Method | Benefits | Impediment | Contribution |
|---|---|---|---|---|---|
| Liu Z et al. [41] | Jul-2021 | v-SVM with neural network, and autoencoder | A lower refinement process descriptor and a powerful Siamese autoencoder based on EEG signals | Small input data | (EEG)-based neuro-feedback |
| Allgaier J et al. [45] | Nov-2021 | Deep learning end-to-end approach | Significant performance | EEG data used might not be enough | EEG features and deep neural network |

### 4.2.2 Model Overview

In Figure 7, we present a comprehensive summary of the work covered in the reviewed articles. The figure showcases the raw EEG signal data, which is then processed using a tool to extract features such as alpha, beta, theta, gamma, and delta. Machine learning algorithms are used to detect and predict whether the extracted EEG data belongs to a person with tinnitus or a healthy individual.

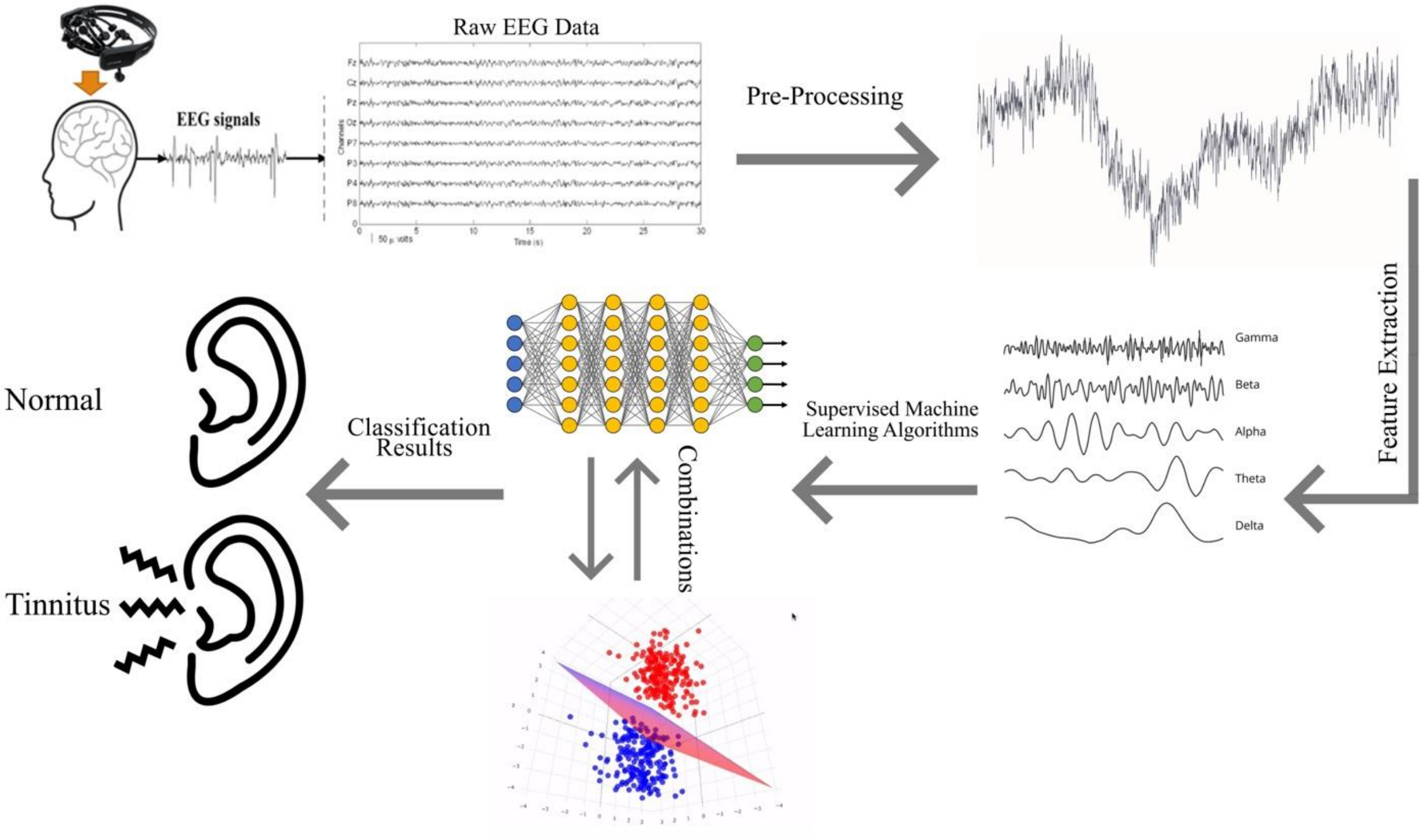

Fig.3. A schematic of machine learning models for Tinnitus diagnosis

### 4.3 Detailed excavation of reviewed studies

Presentation of useful and effective summaries of all the studies that were reviewed in the previous sections on the diagnosis and prediction of tinnitus are given in subsections 4.1 and 4.2. These sections plan to provide a complete analysis of the articles by recommending a five-step method. Almost all the reviewed studies have the same opinion about having five general steps to achieve the goal of tinnitus detection or prediction with EEG signals by machine learning or deep learning. According to

In Figure 4, there are five main steps involved in capturing and collecting EEG signals, removing artifacts and noise signals, performing pre-processing operations, extracting necessary features, and classifying the output. For a detailed description of each step of the methods used in each research, refer to subsections 4.3.1 to 4.3.5. Because all this information is taken from a reliable source, we can compare the results of the studies at each stage at the end.

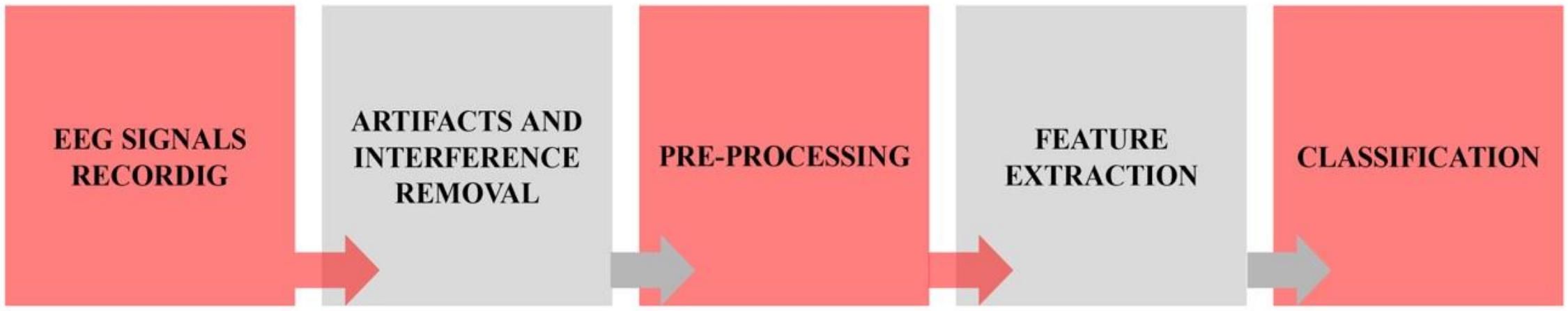

Fig.4. Five common phases to attain the aim of tinnitus detection by ML application EEG signals

### 4.3.1 Resolution of application EEG datasets to assessment tinnitus by ML

Detailed content about the data used in the articles reviewed in the previous sections is given in Table 4. Presenting this data in a table helps us to have a fairly comprehensive set of data, which can significantly affect the choices of future data of researchers. Most of the data in this table includes signals collected from tinnitus patients and healthy people. Since the machine learning algorithm requires significant amounts of data to measure performance ‹the limited number of patients raises concerns about the accuracy of the results.

Table 4. EEG data information

| Location of Data Complex | Ref | Member | Tinnitus | Normal No. | Members Age | Status | Record Duration | Frequency | Channels No. |
|---|---|---|---|---|---|---|---|---|---|
| Department of Otolaryngology, Head and Neck Surgery of the PLA General Hospital | [39] | 20 | 10 | 10 | 18-65 | - | 5 Min | 1000 Hz | 64 |
| | [40] | 42 | 32 | 10 | 20-50 | EC | 10 MIN | - | - |
| Antwerp University Hospital | [38] | 271 | 129 | 142 | - | EC | 5 Min | 128 Hz | 19 |
| Department of Otolaryngology, Sun Yat-sen Memorial Hospital, Sun | [37] | 267 | 187 | 80 | 17-43 | EO | 7min | 1000 HZ | 128 |
| | [42] | 36 | 18 | 18 | 20-63 | EO | 7 min | 1000 HZ | 129 |
| | [43] | 26 | 13 | 13 | - | - | - | 1000 HZ | 64 |
| Yat-sen University | | | | | | | | | |
| Institutional Review Board of the Hallym University College of Medicine | [32] | 22 | 11 | 11 | - | - | - | 1000 HZ | 32 |
| Changhai Hospital in Shanghai | [41] | 55 | 40 | 15 | - | - | - | 50.0 kHz | 1 |
| National Research Foundation of Korea (NRF) | [44] | 417 | 264 | 153 | 20-75 | EC | 5min | 128 HZ | 19 |
| Brain Products Gmbh, Gilching, Germany | [45] | 42 | 29 | 13 | - | - | - | 500 HZ | 62 |
| Iranian National Brain Mapping Laboratory (NBML) | [36] | 16 | 8 | 8 | 25-59 | EO | 5 Min | 1200 Hz | 64 |

### 4.3.2 Resolution of interposition and artifact removal procedures

To accomplish pre-processing actions, EEG signals must not have any noise. The methods used in each article to remove artifacts are listed in Tables 5 and 6. Classifying in table format helps us understand which method is used in most articles. For example, a 50Hz notch filter has been used in three studies.

Table 5. Detail data of accepted interference omission methods

| **Interference Type** | **Method of Removing Interference** | **Research** |
|---|---|---|
| Power Line | A gap filter of 50 Hz | 36, 44,37 |
| Low-Frequency Noise | Band-pass filter (500 Hz) | 45 |
| Irrelevant Signals | Low-pass filtered (100 Hz) | 41 |
| EOG | Band-pass filtered (0.5-70 Hz) | 32 |
| EGI | Band-pass filtered (0.5-80 Hz) | 42 |
| 0.5 Hz and 80 Hz Frequencies | A gap filter of 50 Hz | 37,39 |
| Frequencies Under 2 Hz | The cut-off frequency of 2 Hz | 36 |
| Unnecessary Signals | FIR filter | 37 |
| Power-To-Power Cross-Frequency | Band pass filtered (2-44 Hz) | 44, 38 |
| Trend Descriptor with Lower Fineness | Band-pass filtered (100-1500 Hz) | 41 |
| Power Spectrum Entropy | Band-pass filtered (0.5-44 Hz) | 43 |
| ECG | Band-pass filter (0.5-90 Hz) | 40,42 |
| 50 Hz Power Frequency | Band-pass filter (0.5-90 Hz) | 40 |

Table 6. Compilation of accepted interference elimination methods data

| **Artifacts Type** | **Artifact Removal Method** | **Research** |
|---|---|---|
| Eye Blinks | Independent Component Analysis (ICA) | 36, 42,38,39, 40, 32,44 |
| Eye Blinks | Manually (an experienced neurophysiologist) | 37, 44,42 |
| Eye Blinks | Wavelet Transform Method | 32 |
| Eye Blinks | EEGLAB software | 36 ,40,42,37 |
| Eye Blinks | WinEEG software | 44 |
| Eye Blinks | Weighted Phase Lag Index (WPLI) | 36 |
| Muscles Movement | ICA | 37,38,44,42 |
| Muscles Movement | Manually (an experienced neurophysiologist) | 44 |
| Power Spectrum | Wavelet Transform Method | 43 |
| Sensor Motions | EEGLAB software | 37 |
| Blind Source Separation | Principal Component Analysis (PCA) | 41 |
| Repairing Bad Sensors and Bad Trials | Autoreject library | 45 |
| Heart Beat | ICA | 36 |
| High Frequency Noise | Filtering | 44 |
| Discrete Frequencies, Harmonics | Notch filtering | 44 |

### 4.3.3 resolution of pre-processing function

To recognize tinnitus through EEG signals, the signals must undergo changes that can be collected and used for feature extraction. Relevant information about tinnitus identification is provided in Table 7 of each article.

Table 7. Data collection of pre-processing actions

| **Research** | **Preprocess Methods** |
|---|---|
| Mohagheghian F et al. [36] | • Divided into 18 trials of 5-second (90-seconds) |
| Wang CD et al. [37] | • Segmented into 2 s slices<br>• FIR |
| | • Creating an asymmetric matrix image using delta, theta, alpha, and beta frequency bands |
| Piarulli A et al. [38] | • Using the Desikan-Killiany atlas<br>• Segmented the brain into 68 regions and assigned time courses by averaging the constituent voxels |
| Jianbiao m et al. [39] | • Upload the electrode coordinates file<br>• Removal of electrodes not connected to the central brain<br>• Segmentation into two-second |
| Li Z et al. [40] | • MATLAB-EEGLAB v14.1.2<br>• Divided into 10-second intervals |
| Liu Z et al. [41] | • z-score normalization |
| Sun ZR et al. [42] | • Segmented in at least 20 2 s-epochs |
| Z S et al. [43] | • Improved Welch method<br>• Normalization |
| Hong ES et al. [32] | • Divisional from 500 ms pre-stimulus to 1,000 ms post-stimulus |
| Allgaier J et al. [45] | • Normalization |
| Vanneste S et al. [44] | • Using z-score for normalization |

### 4.3.4 Analyzing the procedure used in the feature extraction phase

The method of extracting input signals by various pieces of research is compiled in Table 8. EEGLAB was used as a feature extraction tool in most studies. Compared to other features, features of frequency and time domain features have been extracted more.

Table 8. Gathering information in the feature extraction feature

| **Research** | **Feature Extraction Method** | **Extracted Features** |
|---|---|---|
| Vanneste S et al. [44] | • sLORETA algorithm | Spatial-temporal features |
| Allgaier J et al. [45] | • Convolutional layer (end-to-end)<br>• Autoreject library | Reduce noise data |
| Mohagheghian F et al. [36] | • MATLAB (EEGLab and Fieldtrip toolboxes)<br>• BCT | • Delta (2–3.5 HZ), Theta (4–5.2Hz), AlphaI (9–10 Hz)<br>• Effective connectivity image between EEG signals |
| Wang CD et al. [37] | • MATLAB R2013a | Alpha1 (8–10 Hz), Alpha2 (10–12 Hz), Delta (2–3.5 Hz), Theta (4–7.5 Hz), B1 (13–18 Hz), B2 (18.5–21 Hz), Beta3 (21.5–30 Hz) and Gamma (30.5–44 Hz) |
| Piarulli A et al. [38] | • OpenMEEG | Gamma (2–3.75 Hz); Beta (4–7.75 Hz); Alpha (8–11.75 Hz); B1 (12–17.75 Hz); B2 (18–29.75 Hz). |
| Jianbiao M et al. [39] | • Combination of Wavelet Packet Transform and Sample Entropy | Analysis of:<br>• Time domain<br>• Frequency domain<br>• Time-frequency<br>• Non-linear kinetic<br><br>Statistical mean feature |
| Li Z et al. [40] | • PLV<br>• PLI<br>• PCC | • Time domain<br>• Frequency domain<br>• Time domain statistical |
| Liu Z et al. [41] | • Encoder<br>• Decoder | • Display with low dimensions<br>• Reconstruction of raw data |
| Sun ZR et al. [42] | • PCA<br>• PCA+FFT | • Time domain<br>• Frequency domain<br>• Time domain statistical |
| Z S et al. [43] | • Wavelet Transform (WT)<br>• db4 Wavelet | • Delta (0.5-3.5Hz), Theta (4-7.5Hz), Alpha (8- 12Hz), Beta (13-30Hz) and Gamma (30.5-44Hz)<br>• wavelet function<br>• decompose EEG signals |
| Hong ES et al. [32] | • Morlet Wavelet Transform | • Time-frequency response |

### 4.3.5 Resolution of ML and DL classifiers for Tinnitus assessment

A summary of the algorithms and models used in the reviewed studies is given in Table 9. The target of this table is to provide an abstract of the algorithms used in diagnosing or predicting tinnitus.

Table 9. Data collection from ML classification

| **Research** | **Publish Year** | **ML/DL Algorithm** | **Classification Performance** | **Performance** |
|---|---|---|---|---|
| Mohagheghian F et al. [36] | 2019 | SVM | LOOCV | Accuracy= 91.7% |
| Wang CD et al. [37] | Jul-23 | MECRL<br>v-SVM<br>MLP<br>EEGNet<br>SiameseAE<br>SMeta-SAE<br>4D-CNN | ACC<br>AUC<br>F1-score<br>Precision<br>Recall | Accuracy= 90.34% |
| Piarulli A et al. [38] | Feb-23 | SVMs | 5-folds cross-validation | Accuracy=94.6% |
| Jianbiao M et al. [39] | 2021 | SVMs | 10/15-fold cross-validation | Accuracy=90.5% |
| Allgaier J et al. [45] | Nov-21 | Keras | Random seeds (NumPy, TensorFlow and the OS environment) | Accuracy=75.6% |
| Vanneste S et al. [44] | 2018 | SVM | K-fold cross-validation (RMSE, TRP, FPR, k-Statistic) | Accuracy=53.30% |
| Li Z et al. [40] | Jan-22 | SVM<br>MLP<br>CNN<br>CNN-LSTM | 10-fold cross-validation<br>Loo-CV | Accuracy=99.42 |
| Liu Z et al. [41] | Jul-21 | v-SVM | ShallowNet<br>DeepNet | Accuracy= 79.17%<br>Accuracy= 88.89% |
| Sun ZR et al. [42] | May-2018 | SVM | Different cross-validation | Accuracy=99.72% |
| Z S et al. [43] | Jul-20 | DNN<br>CNN<br>RNN | ReLU | Accuracy=92% |
| Hong ES et al. [32] | Jul-20 | CNN<br>EEGNet<br>SVM | 11-fold cross-validation leaving a pair | Accuracy= 0.774<br>Accuracy= 0.759 |

## 5. Discussion and comparison

Taking into account the adopted method of SLR, the previous sections were in terms of providing a comprehensive explanation of the article selection method along with a summary of the final studies to achieve their essence and a detailed analysis of all research reviewed based on the defined five-step model. In this section, the questions presented in the third part have been fully answered.

AQ1: Which machine learning algorithms have been used to diagnose or predict tinnitus?

Fig. 4 illustrates various machine learning algorithms used for diagnosing or predicting tinnitus in subjects based on EEG signals. In Section 4, we designed a taxonomy and classified the algorithms accordingly. SVM has been frequently used for prediction purposes. Two machine learning structures have been used in most of the papers and they were tested with a similar data set to ensure their accuracy. SVM and DNN convolutional block models outperform other methods as they provide consistent results in multiple tests of the SVM architecture and excellent predictive performance with robustness to the overfitting of DNN structures. Based on the reviewed studies, it can be concluded that combining external methods with machine learning structures provides significant accuracy.

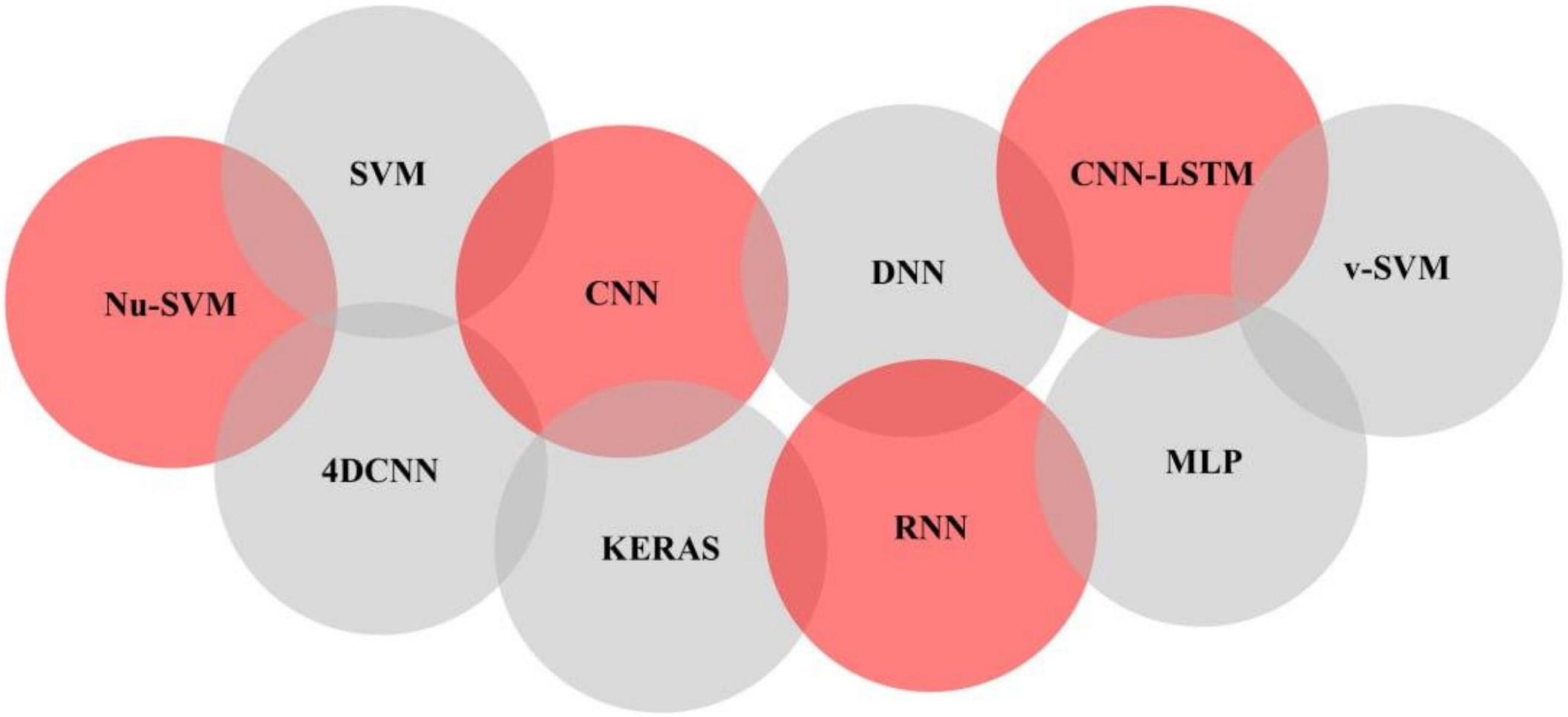

Fig. 5. Diversity of adopted ML and DL algorithms in the reviewed research

AQ2: Which machine learning methods are most preferable to achieve our purpose?

The distribution of ML architectures used in the surveyed research is presented in Figure 6. The figure indicates that approximately half of the implemented ML classifiers were based on support vector machines (SVM), including v-SVM, SVM-LooCV, and SVM-10CV, and convolutional neural network (CNN) based methods such as CNN, 1DCNN, 2DCNN, and 3DCNN, which were preferred over other algorithms. SVM was the most commonly used ML structure, accounting for 72.72% of the total percentage, to diagnose or predict tinnitus.

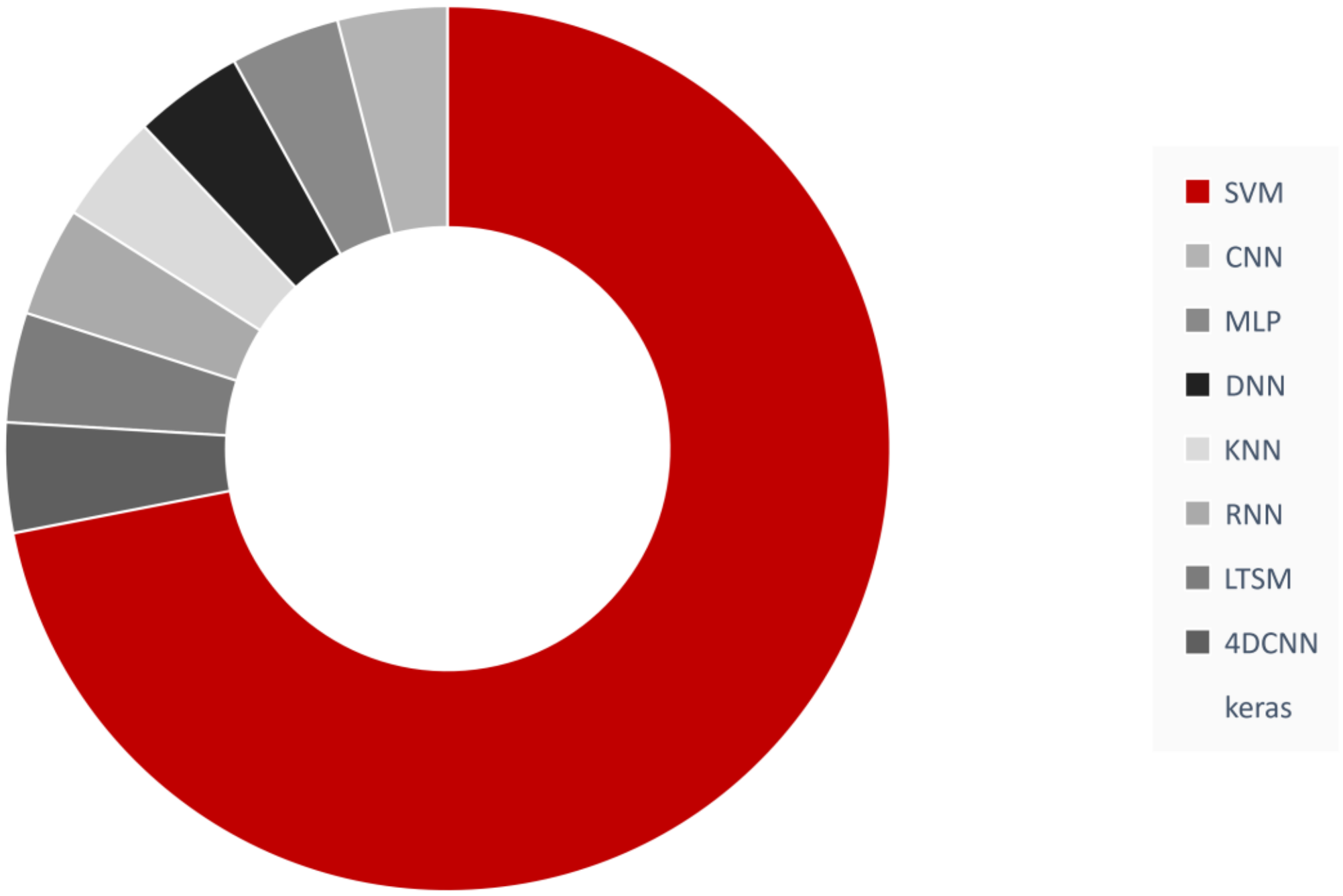

Fig.6. Distribution of adopted ML and DL across all studied papers

AQ3: What are the primary methods utilized for extracting features from EEG signals?

In Figure 7, you can see the various techniques that were used to obtain the specific characteristics of the recorded measurements. For instance, MATLAB was used as a tool for functional performance, as shown in the figure. The researchers also discovered that some of them utilized two or more techniques from the ones presented in Figure 7 to generate different combinations of features.

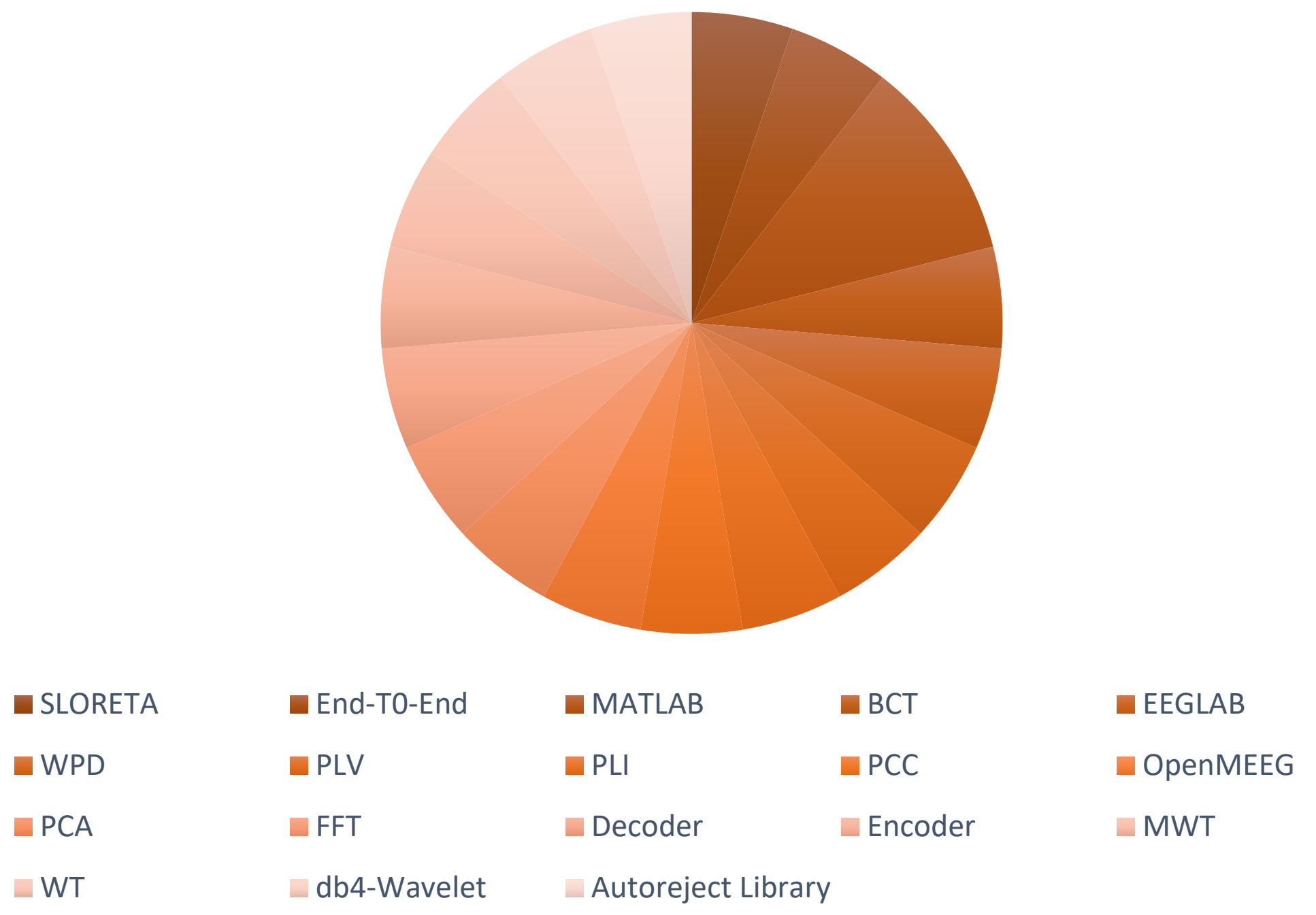

Fig.7. Frequencies of adopted feature extraction methods

### 5.2 Open obstacle and future works

Based on the SLR method, this article analyzes the conducted research and the main features, including the positive and negative ones. Despite these analyses, some uncertainties and challenges still exist and need to be addressed. In this section, we will present these challenges in detail.

AQ4: What are the future and open works relevant to tinnitus diagnosis or prediction?

Through this part, we discuss the problems that researchers have computed during the implementation of the five-step methods, such as the source of input data, the implementation of ML and DL models, and noise removal and data preprocessing, in this section. Additionally, we will examine the overall issue of predicting tinnitus and its significance in resolving it.

Insufficient input data: One major challenge faced by researchers in the detection or prediction process was the lack of input data. This led to overfitting and unreliable accuracy. Some studies attempted to dominate these issues by implementing techniques like exploiting dropout layers with different probability rates to deal with overfitting. To crop more data, the method of windowing and generating samples was used. Although a lot of work was done, the lack of input data is one of the important challenges that must be addressed in future research.

Implementing and usability: One of the key problems that researchers faced in their studies was the implementation of their proposed methods. Although providing scientific and practical methods to help clinical environments is one of the major purposes of research, the models introduced by some researchers did not achieve this basic goal. One of the basic obstacles to using these methods to evaluate tinnitus in real situations was the selection of deep learning models with many layers to attain high accuracy. Additionally, some studies recorded EEG signals with many electrodes, such as 19-channel signals, making it challenging to collect these high-channel signals in clinical settings. Therefore, it is essential to focus on designing models that are compatible with the clinical setting to address this issue.

Lack of peculiar and impressive methods for data pre-processing: Data preparation for prediction or detection was a common challenge among most works. Some manually accomplished this task, while others used experience-based methods listed in Tables 5 and 6 for noise removal. It's important to note that pre-processing steps, such as utilizing time window segmentation of different lengths, the AEP method, or transforming signals to images, can negatively impact the precision of the model. These experimental techniques may lead to the removal of important features of the input signal and loss of non-interval exclusivity of signals when alteration to images. Therefore, it's crucial to use a particular and reliable method of data preparation to ensure accuracy.

Giving less attention to tinnitus prediction: During our search for papers on predicting tinnitus, we found very few studies on the subject. We could only review one article that met our selection criteria. Tinnitus can have a serious impact on an individual's well-being, and may even be life-threatening. It can also cause financial strain and put pressure on the healthcare system and economy. Additionally, depression is often diagnosed late in tinnitus sufferers, which can make treatment difficult and sometimes even impossible. Given these factors, more time and attention must be devoted to researching the prediction of this chronic and debilitating condition.

## 6. Conclusion

The focus of this paper is the diagnosis and prediction of tinnitus through EEG signals with the aid of machine learning algorithms. The study utilized the SLR method to conduct a comprehensive review of relevant literature and analyze their key aspects. Furthermore, the paper discussed the topics that need to be addressed in future studies. Based on our objective and the common practice in research articles, we compare the performance of different machine learning and deep learning algorithms on a given dataset. The classification is based on a set of all machine learning techniques used across various research projects. Analyzing 11 SLR-based articles, we concluded that SVM-based machine learning methods namely v-SVM, nv-SVM, and SVM-based AUC were preferred among the different algorithms used, with approximately 72.72% of the total. It was consummate that the deep learning methods relying on CNN, i.e. CNN, 4DCNN, and CNN-LSTM have occupied the second place. Most studies used the convolutional FFT technique to extract local features, although different feature extraction methods were used by different researchers. Most studies analyzing tinnitus use a similar method: removing noise and artifacts, collecting EEG signals, extracting essential features from pre-processed signals, and classifying tinnitus and normal subjects using one or more machine learning methods. In summary, our goal in conducting this research was to provide a comprehensive review of subsequent research with a strong foundation in this field, using the SLR method.

## Abbreviations

| | |
|---|---|
| **4DCNN** | Four-Dimensional Convolution Neural Network |
| **ACC** | Accuracy |
| **AUC** | Area Under the Curve |
| **BCT** | Brain Connectivity Toolbox |
| **CNN** | Convolutional Neural Network |
| **DL** | Deep Learning |
| **DNN** | Default Neural Network |
| **EC** | Eye Closed |
| **ECG** | Components Of the Electrocardiogram |
| **EEG** | Electroencephalogram |
| **EO** | Eye Open |
| **EGI** | Electrical Geodesics Incorporated |
| **EOG** | Electrooculography |
| **FFT** | Fast Fourier Transform |
| **fMRI** | Functional Magnetic Resonance Imaging |
| **FPR** | False-Positive Ratio |
| **ICA** | Independent Component Analysis |
| **ISI** | International Statistical Institute |
| **LSTM** | Long Short-Term Memory |
| **LOOCV** | Leave-One-Out Cross-Validation |
| **MEG** | Magnetoencephalography |
| **ML** | Millilambert |
| **MLP** | Multilayer Perceptron |
| **MSE** | Mean Square Error |
| **NO** | Number |
| **PCA** | Principal Component Analysis |
| **PLI** | Phase Lag Index |
| **PLV** | Posterior Left Ventricular |
| **ROC** | Receiver Operating Characteristic |
| **ROI** | Return On Investment |
| **SNN** | Spiking Neural Networks |
| **SLR** | Systematic Literature Review |
| **SVM** | Support Vector Machine |
| **TFI** | Tinnitus Functional Index |
| **TPR** | True-Positive Ratio |
| **WPLI** | Weighted Phase Lag Index |

**Ethical standards followed:**

The authors involved in this study declare that they do not have any conflicts of interest to report.

**Informed consent:**

Not applicable.

**Source of Fund**

No funding agency, public sector, or business was involved in this research.

## Authors Biography

**Farzaneh Ramezani**

BA in clinical psychology from Isfahan University, Isfahan, Iran. 2011-2015

Master's degree in clinical psychology from Tabriz University, Tabriz, Iran. 2016-2018

Fields of Interest: Neuroscience, Artificial Intelligence, and Psychology

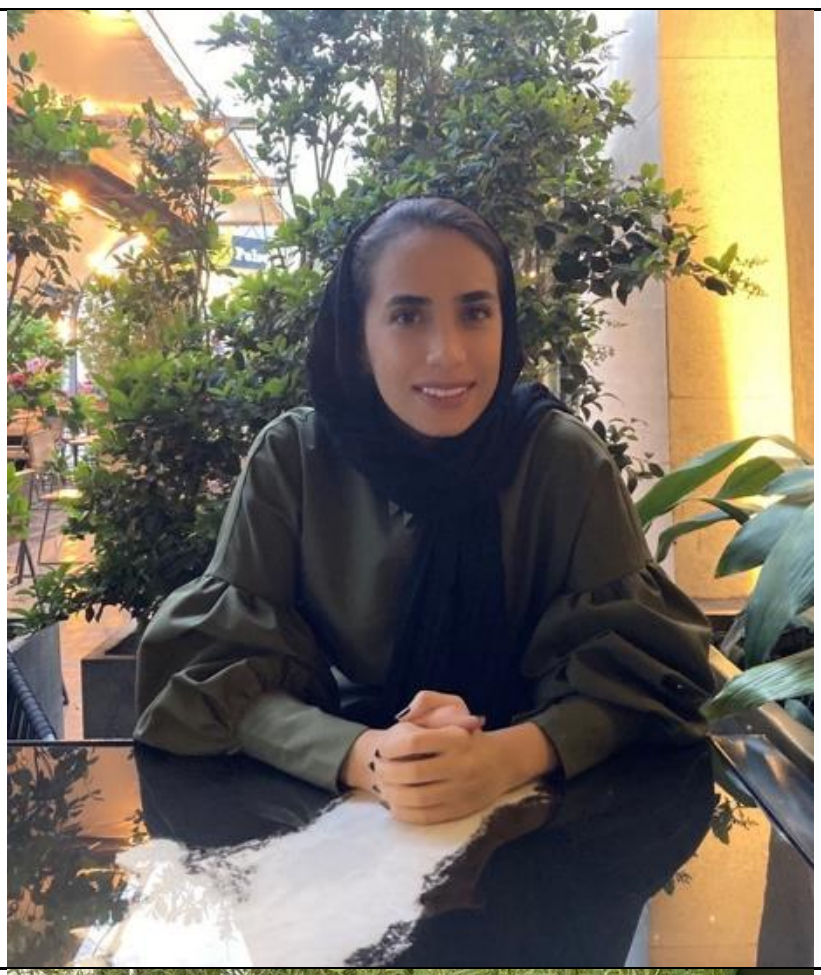

**Hamidreza Bolhasani, PhD**

AI/ML Researcher / Visiting Professor

Founder and Chief Data Scientist at DataBiox

Ph.D. Computer Engineering from Science and Research Branch, Islamic Azad University, Tehran, Iran. 2018-2023.

Fields of Interest: Machine Learning, Deep Learning, Neural Networks, Computer Architecture, Bioinformatics

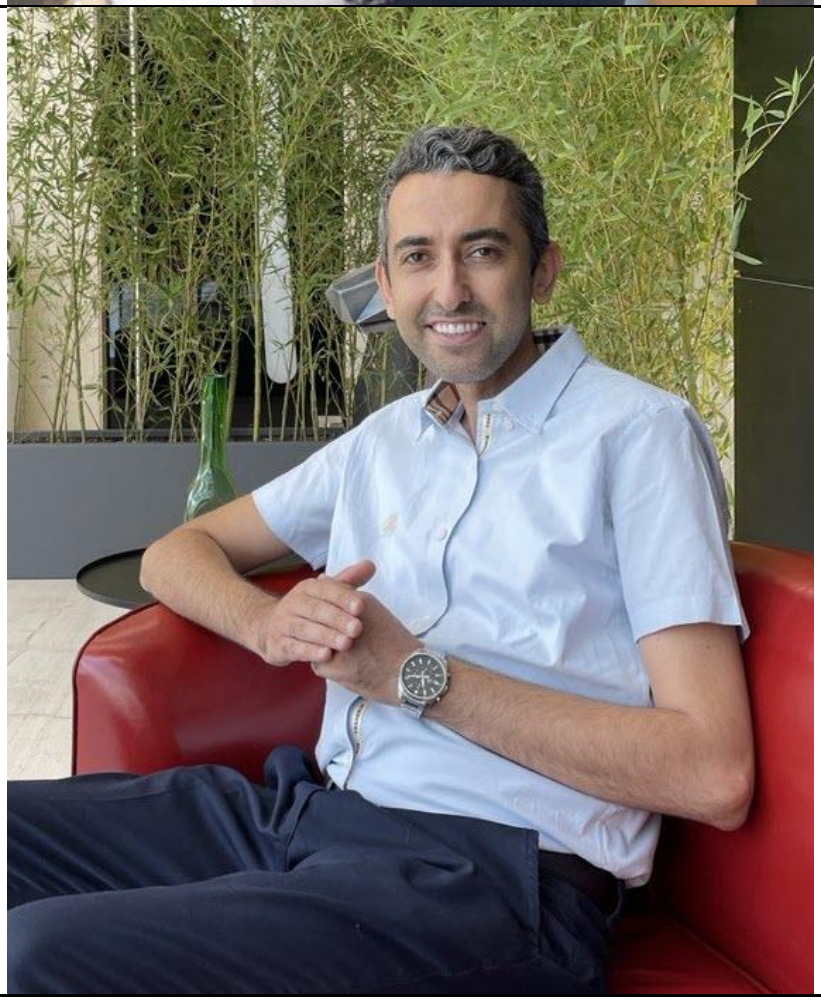